\documentclass[lettersize,journal]{IEEEtran}
\usepackage{amsmath,amsfonts}
\usepackage{algorithmic}
\usepackage{algorithm}
\usepackage{array}
\usepackage[caption=false,font=normalsize,labelfont=sf,textfont=sf]{subfig}
\usepackage{textcomp}
\usepackage{stfloats}
\usepackage{url}
\usepackage{verbatim}
\usepackage{graphicx}
\usepackage{cite}
\usepackage{dsfont}
\begin{document}

\title{EventSSEG: Event-driven Self-Supervised Segmentation with Probabilistic Attention}


\author{Lakshmi Annamalai and Chetan Singh Thakur
\thanks{Lakshmi Annamalai is with Defence Research and Development Organization, Bangalore and Indian Institute of Science, Bangalore.}
\thanks{Chetan Singh Thakur is with Department of Electronics Systems Engineering, Indian Institute of Science, Bangalore.}}



\maketitle

\begin{abstract}
Road segmentation is pivotal for autonomous vehicles, yet achieving low-latency and low-compute solutions using frame-based cameras remains a challenge. Event cameras offer a promising alternative. To leverage their low-power sensing, we introduce EventSSEG, a method for road segmentation that uses event-only computing and a probabilistic attention mechanism. Event-only computing poses a challenge in transferring pre-trained weights from the conventional camera domain, requiring abundant labeled data, which is scarce. To overcome this, EventSSEG employs event-based self-supervised learning, eliminating the need for extensive labeled data. Experiments on DSEC-Semantic and DDD17 show that EventSSEG achieves state-of-the-art performance with minimal labeled events. This approach maximizes event cameras' capabilities and addresses the lack of labeled events.
\end{abstract}


\section{Introduction}
\IEEEPARstart{I}{n} the realm of autonomous driving, road segmentation is a fundamental task for visual understanding and navigation \cite{Zhang}. Current methodologies primarily rely on conventional frame-based cameras \cite{road_new_5} \cite{road_3} \cite{road_5}. However, these systems present computational complexity and time consumption owing to the dense and synchronous nature of data, which contradicts the real-time, computationally constrained requirements of autonomous driving. 

The breakthrough in the endeavour to mimic biological vision has paved the way for the invention of a novel sensor known as an event camera \cite{camera_5}. In contrast to conventional cameras, event cameras trigger pixels independently upon changes in intensity. Consequently, the event camera generates sparse and asynchronous data, thereby granting advantages such as low latency and low compute processing. With its potential to overcome the limitations of frame-based cameras, this paradigm shift holds promise for the resource-constrained robotics community.

Existing event-based segmentation methods stack events into synchronous frames to make them compatible with synchronous 2D vision networks. This transformation introduces unwarranted latency and fails to preserve the sparsity of events. Hence, a redesign of neural networks is required to effectively integrate event data. Redesigning neural networks will render pre-trained models of conventional vision networks incompatible with the event domain. This will lead to the requirement for a substantial amount of labelled event data. Since event modality is a relatively emerging field compared to its frame-based counterpart, annotated data is still scarce in the event domain. Moreover, event-level labelling is more labour-intensive than pixel-level labelling. 

How can we circumvent the scarcity of labelled event data while working with a data-hungry, event-driven deep neural network? With this question in mind, this paper puts forth EventSSEG (Event-driven Self-Supervised Segmentation network), a novel event-only transformer-based road segmentation network that uses raw event data and self-supervised learning. 

A key innovation presented in this work is the application of self-supervised learning to event point cloud data. In contrast to the existing approaches, the proposed self-supervised task is defined on raw event point cloud data. Our event-based self-supervised framework involves two key stages: i) Training the EventSSEG to solve a pretext task on unlabeled events and ii) Fine-tuning the network on downstream road segmentation tasks with very few labelled events. 

\section{Related Work}
This section provides an in-depth survey of two related domains of road segmentation: semantic segmentation and self-supervised learning using an event camera.

\subsection{Semantic Segmentation in Event Domain}
The first baseline for semantic segmentation using an event camera was presented by Ev-SegNet \cite{evsegnet}. An Xception-based CNN \cite{xception} was built on a novel $6$ channel event representation. Additionally, \cite{evsegnet}, an extension of DDD17 dataset \cite{ddd17} designed for semantic segmentation task, was introduced by \cite{evsegnet}. 

\cite{v2e} converted video data to synthetic events in order to generate more data for event-based semantic segmentation. Following this, \cite{evdistill} proposed a novel approach to train a student network on unlabeled event data by distilling knowledge from a teacher network. Using both frame-based and event-based data, \cite{isaffe} proposed a novel event-based multi-modal semantic segmentation framework. ESS \cite{ess}, in order to leverage unpaired events and frames, proposed unsupervised domain adaptation. \cite{Jia} introduced transformer architecture with a posterior attention mechanism. 

The method in \cite{review_2} aims to reduce latency with parallel multi-latent memories operating at varying rates, but this results in high memory usage and bandwidth demands, along with memory fragmentation issues. OpenESS proposed a multi-modal approach \cite{review_1} that combines information from image, text, and event data, but our method focuses exclusively on event-based segmentation, which allows us to fully leverage the unique properties of event data.

\subsection{Self Supervised Learning in Event Domain}
The solution to reducing the dependencies on large-scale labelled datasets is to adopt self-supervised learning (SSL) \cite{ss_vision_1} \cite{ss_vision_2} \cite{ss_vision_7}.

The absence of labelled event data in a number of vision tasks such as optical flow \cite{ss_1}, intensity reconstruction \cite{ss_2}, object classification \cite{ss_3} \textit{etc.}, has been addressed by SSL. These works either rely on massive amounts of RGB data or synchronous, pixel-aligned RGB and event data recording. \cite{evtransfer} \cite{ess} \cite{evdistill} transfer knowledge from the RGB domain to the event domain with unpaired labelled frames and unlabeled events. These methods rely on data acquired under similar conditions from both frame-based and event-based cameras, which is often not feasible. Masked Event Modeling (MEM), an SSL framework dependant only on event data, was proposed \cite{mask}. It employs partially removed event data reconstruction as a pre-task. However, ours is the first work to propose an SSL framework for raw event data with a pre-task constructed around event polarity. 

\section{Proposed Solution}

\begin{figure*}[tb]
\centering
\subfloat[Self-Supervised EventSSEG]{\includegraphics[width=0.4\linewidth,keepaspectratio=true]{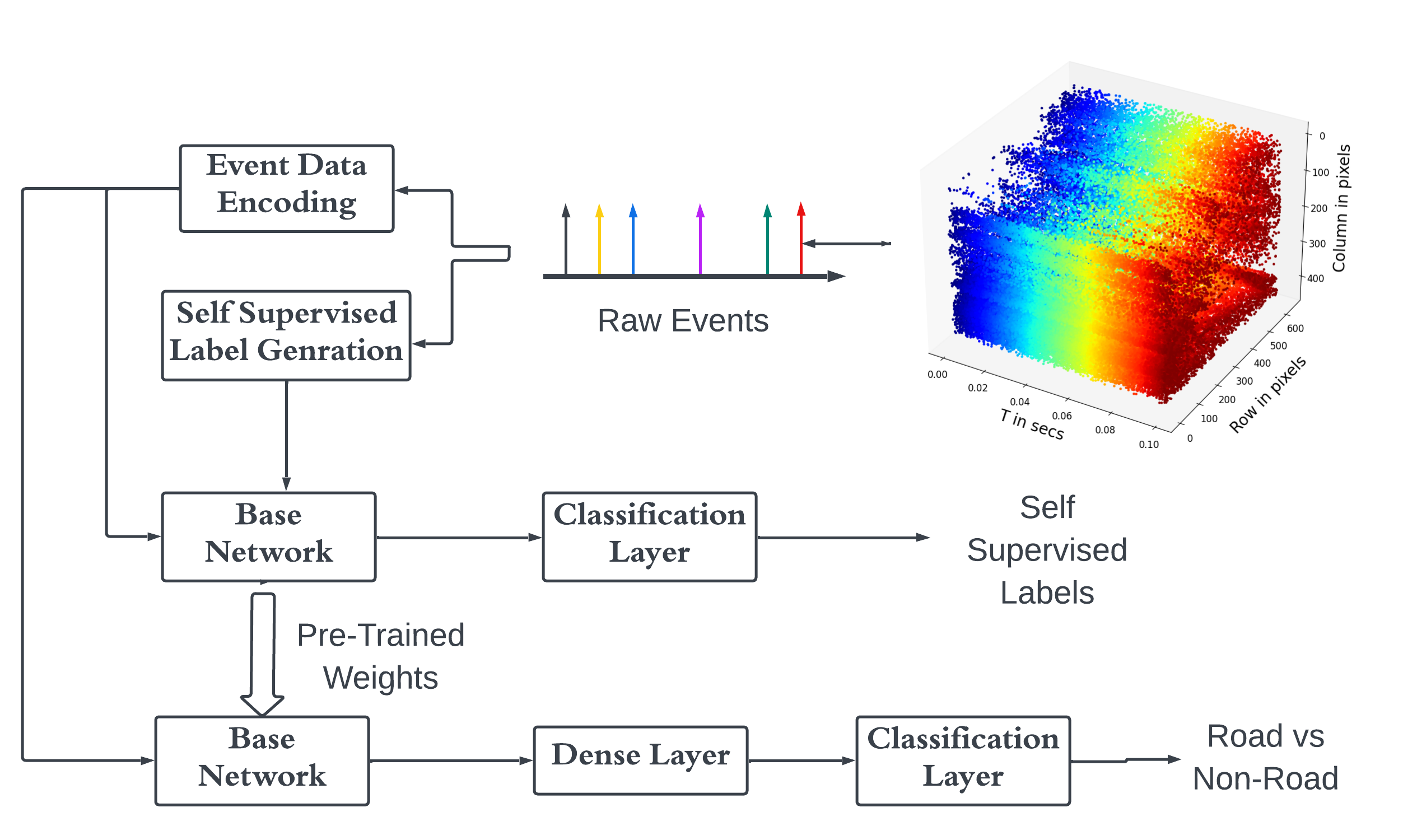}} 
\subfloat[Base Network of EventSSEG]{\includegraphics[width=0.4\linewidth,keepaspectratio=true]{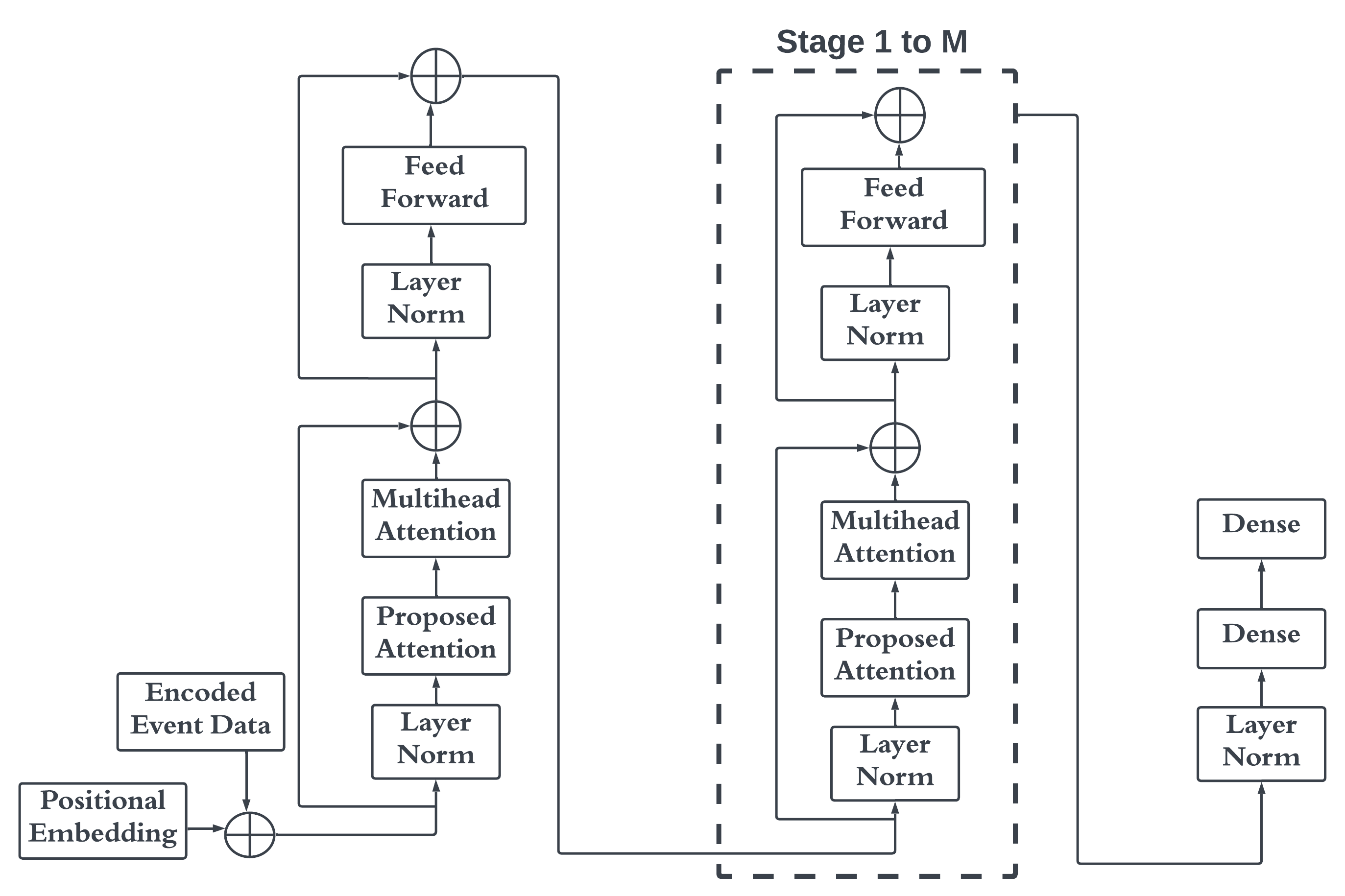}}
\caption{(a): Framework of EventSSEG for self-supervised learning of pre-trained weights. Self-supervised labels are generated from the polarity of raw events. A base network and a classification layer are trained to minimize the cross-entropy loss between true and predicted self-supervised labels. Once training converges, the classification layer is replaced with a small feed-forward neural network with a dense layer of $128$ neurons and a classification layer. The network is fine-tuned with a few labelled samples of the downstream road segmentation task. (b) Base network of EventSSEG. Transformer architecture with four feature extraction blocks. Each block comprises layer normalization, residual connection, multi-head attention with the proposed attention mechanism and a feed-forward neural network.}
\label{fig:qualitative}
\end{figure*}  

We describe the EventSSEG engineered to handle raw event stream. Drawing inspiration from the self-supervised learning algorithms in conventional vision, the proposed architecture learns pre-trained weights through a self-supervised pre-task defined on raw events.

\subsection{Event Camera Preliminary}
The change of log-scale intensity at pixel location $(x,y)$ above a pre-defined threshold at time $t$ triggers an event represented as a quadruple $e=(x,y,t,p)$, where $p\in{[+1,-1]}$ represents increase or decrease in brightness respectively. 

\subsection{Attention in Transformers}

This section brings out the general attention mechanism followed in Transformer architecture \cite{transformer}. The given input sequence $E=\left[e_1,e_2\ldots{e_N}\right]$ $\in$ $R^{N\times{d_e}}$ is projected into query $\left(Q\in{R^{N\times{d}}}\right)$, key $\left(K\in{R^{N\times{d}}}\right)$ and value $\left(V\in{R^{N\times{d}}}\right)$ matrices through the following transformation,

\begin{equation}
Q=EW^T_q;K=EW^T_k;V=EW^T_v
\end{equation}

Where, $W_q,W_k$ and $W_v$ $\in{R^{d\times{d_e}}}$ are the weight matrices, $Q=\left[q_1,q_2\ldots{q_N}\right]$, $K=\left[k_1,k_2\ldots{k_N}\right]$ and $V=\left[v_1,v_2\ldots{v_N}\right]$ and $q_i,k_i$ and $v_i$ are query, key and value vectors respectively. 

The $i^{th}$ output is estimated as the weighted average of the value vectors, $o_i=\sum_{j=1}^{N}w_{ij}v_j$, where the attention scores $w_{ij}$ are generally estimated as scaled dot product attention as follows,

\begin{equation}
softmax\left(\frac{q^T_ik_j}{\sqrt{d}}\right)
\end{equation}

\subsection{Probabilistic Attention Estimation}

In this work, we propose a novel attention score inspired by \cite{prob_keys}. The aim of the proposed attention mechanism is to incorporate spatial information of event occurence and to introduce probability in the relation between key and query.

\begin{equation}
w_{ij}=\mathds{P}(k_j|q_i)+\mathds{P}(\Delta_j|q_i) 
\label{total_prob}
\end{equation}

Where $\Delta_j=|x_i-x_j|+|y_i-y_j|$ is the spatial distance between event $i$ and $j$ and 

\begin{enumerate}
\item[*]{$\mathds{P}(k_j|q_i)$ is the posterior probability of getting key vector $k_j$ given the query $q_i$}
\item[*]{$\mathds{P}(\Delta_j|q_i)$ is the posterior probability of event $j$ occuring at a spatial distance of $\Delta_j$ from the current event $i$, given the query $q_i$.}
\end{enumerate}

\subsubsection{First Term of Eq. \ref{total_prob}}

We start with Bayes' theorem to express $\mathds{P}(k_j|q_i)$:

\begin{equation}
\mathds{P}(k_j|q_i)=\frac{\mathds{P}(q_i|k_j)\mathds{P}(k_j)}{\mathds{P}(q_i)}
\label{eq:bayes}
\end{equation}

Following distributions are assumed for $\mathds{P}(q_i|k_j)$, $\mathds{P}(k_j)$ and $\mathds{P}(q_i)$,

\begin{enumerate}
\item[*]{$\mathds{P}(q_i|k_j)\sim{\mathcal{N}(q_i|k_j,\sigma_j^2\mathbf{I})}$, where $\mathbf{I}$ is identity matrix}
\item[*]{$\mathds{P}(k_j)=\pi_j$}
\item[*]{$\mathds{P}(q_i)=\sum_{k=1}^{N}\mathds{P}(q_i|k_k,\Delta_k)\mathds{P}(k_k,\Delta_k)$, a Gaussian mixture model with $\mathds{P}(k_k,\Delta_k)=\gamma_k$ and $\mathds{P}(q_i|k_k,\Delta_k){\sim}\mathcal{N}(q_i|k_k,\sigma_{q_k}^2\mathbf{I})$}
\end{enumerate}

Substituting the above into Eq. \ref{eq:bayes},

\begin{equation}
\mathds{P}(k_j|q_i)=\frac{\frac{\pi_j}{\sigma_j}\text{exp}\left[-\frac{(q_i-k_j)^T(q_i-k_j)}{2\sigma_j^2}\right]}{\sum_{k=1}^{N}\frac{\gamma_k}{\sigma_{q_k}}\text{exp}\left[-\frac{(q_i-k_k)^T(q_i-k_k)}{2\sigma^2_{q_k}}\right]}
\end{equation}

Simplifying, we get,

\begin{equation}
\frac{\frac{\pi_j}{\sigma_j}\text{exp}\left[-\frac{\left(\Vert{q_i}^2\Vert+\Vert{k_j}^2\Vert\right)}{2\sigma_j^2}\right]\text{exp}\left(\frac{q_i^Tk_j}{\sigma_j^2}\right)}{\sum_{k=1}^{N}\frac{\gamma_k}{\sigma_{q_k}}\text{exp}\left[-\frac{\left(\Vert{q_i}^2\Vert+\Vert{k_k}^2\Vert\right)}{2\sigma_{q_k}^2}\right]\text{exp}\left(\frac{q_i^Tk_k}{\sigma_{q_k}^2}\right)}
\end{equation}

Assuming $q_i$ and $k_j$ are normalized, $\mathds{P}(k_j|q_i)$ turns out to be:

\begin{equation}
\frac{\frac{\pi_j}{\sigma_j}\text{exp}\left[-\frac{1}{\sigma_j^2}\right]\text{exp}\left(\frac{q_i^Tk_j}{\sigma_j^2}\right)}{\sum_{k=1}^{N}\frac{\gamma_k}{\sigma_{q_k}}\text{exp}\left[-\frac{1}{\sigma_{q_k}^2}\right]\text{exp}\left(\frac{q_i^Tk_k}{\sigma_{q_k}^2}\right)}
\label{eq:first}
\end{equation}

\subsubsection{Second Term of Eq. \ref{total_prob}}

Using Bayes' theorem again:

\begin{equation}
\mathds{P}(\Delta_j|q_i)=\frac{\mathds{P}(q_i|\Delta_j)\mathds{P}(\Delta_j)}{\mathds{P}(q_i)}
\label{eq:delta}
\end{equation}

We model $\mathds{P}(q_i|\Delta_j)$ and $\mathds{P}(\Delta_j)$ as follows,

\begin{enumerate}
\item[*]{$\mathds{P}(q_i|\Delta_j)\sim{\mathcal{N}(q_i|\Delta_jq_i,\sigma_{\Delta_j}^2\mathbf{I})}$. This makes sure that it utilizes the information obtained from events that occur spatially far from the current event $e_i$.}
\item[*]{$\mathds{P}(\Delta_j)=\beta_j$}
\end{enumerate}

Substituting the above into Eq. \ref{eq:delta}, we get,

\begin{equation}
\mathds{P}(\Delta_j|q_i)=\frac{\frac{\beta_j}{\sigma_{\Delta_j}}\text{exp}\left[-\frac{(q_i-\Delta_jq_i)^T(q_i-\Delta_jq_i)}{2\sigma_{\Delta_j}^2}\right]}{\sum_{k=1}^{N}\frac{\gamma_k}{\sigma_{q_k}}\text{exp}\left[-\frac{(q_i-k_k)^T(q_i-k_k)}{2\sigma^2_{q_k}}\right]}
\end{equation}

Reducing the terms $(q_i-\Delta_jq_i)^T(q_i-\Delta_jq_i)$ and $(q_i-k_k)^T(q_i-k_k)$ and incorporating the assumption that $q_i$ is normalized, $\mathds{P}(\Delta_j|q_i)$ becomes,

\begin{equation}
\frac{\frac{\beta_j}{\sigma_{\Delta_j}}\text{exp}\left[-\frac{\left(1+\Delta_j^2\right)}{2\sigma_{\Delta_j}^2}\right]\text{exp}\left(\frac{\Delta_j}{\sigma_{\Delta_j}^2}\right)}{\sum_{k=1}^{N}\frac{\gamma_k}{\sigma_{q_k}}\text{exp}\left[-\frac{1}{\sigma_{q_k}^2}\right]\text{exp}\left(\frac{q_i^Tk_k}{\sigma_{q_k}^2}\right)}
\label{eq:second}
\end{equation}

\subsubsection{Combining First and Second Term}

Substituting Eq. \ref{eq:first} and \ref{eq:second} into Eq. \ref{total_prob}, we get,

\begin{equation}
w_{ij}=\frac{\frac{\pi_j}{\sigma_j}\text{exp}\left[-\frac{(1-q_i^Tk_j)}{\sigma_j^2}\right]+\frac{\beta_j}{\sigma_{\Delta_j}}\text{exp}\left[\frac{-\left(1-2\Delta_j+\Delta_j^2\right)}{2\sigma_{\Delta_j}^2}\right]}{\sum_{k=1}^{N}\frac{\gamma_k}{\sigma_{q_k}}\text{exp}\left[-\frac{1}{\sigma_{q_k}^2}\right]\text{exp}\left(\frac{q_i^Tk_k}{\sigma_{q_k}^2}\right)}
\end{equation}

It could be seen that i) $w_{ij}$ is directly proportional to the dot product of $q_i$ with respect to $k_j$ and ii) the higher the value of $\Delta_j$, the higher the value of $w_{ij}$.

\subsection{Event Transformer with Probabilistic Attention}

The proposed architecture starts with an input tensor $E=\{e_1,e_2\ldots{e_N}\}$ of dimension $N\times{4}$, where $N$ represents the number of events and $e_i=(x_i,y_i,t_i,p_i)$. A fully connected layer projects the input tensor onto vectors of dimension $12$. Positional embeddings have been added via the standard positional embedding layer. Subsequently, the architecture incorporates four feature extraction blocks, each of which consists of a multi-head attention and a feed-forward neural network. Multi-head attention employs $4$ heads with the proposed probabilistic attention mechanism. The output of multiple heads is combined later. The feed-forward neural network includes two dense layers with $24$ and $12$ nodes with GeLU activation. To ensure stability and accelerate convergence, layer normalization and residual connections are integrated. The feature extraction is followed by a feed-forward neural network with $2048$ and $1024$ neurons with GeLU activation. This base architecture is appended with task-specific layers.

\subsection{Self Supervised Learning Framework}
\label{Self_supervised}
Transformer networks are typically trained using data paired with their corresponding supervisory labels. In the conventional vision, state-of-the-art transformer networks are learnt from a huge corpus of supervisory images. Unfortunately, being an emerging field, the event camera domain suffers from a lack of sufficient labelled data pertaining to specific tasks. Hence, to obtain better performance with limited supervisory signals, the proposed approach investigates the utility of the information embedded in the polarity of the event sequence as a supervisory signal. While trained with the proposed polarity-based self-supervised labels, EventSSEG arrives at semantically meaningful feature representations. These representations can be fine-tuned to the given task with minimal labelled samples.

Given an unlabeled training dataset $E=\{e_i\}_{i=1}^{N}$ of $N$ events, the distribution of polarity is modelled as binomial distribution, and the probability of occurrence of positive events is estimated as 

\begin{equation}
p^+=\sum_{i=1}{N}\mathds{I}(p_i==1)
\end{equation}

. The entropy of the event sequence $E$ is estimated as follows,

\begin{equation}
H = -p^+\text{log}p^+-(1-p^+)\text{log}(1-p^+)
\end{equation}

The supervisory signal $y^s_i$ for each event sequence $E_i$ is formed as follows,

\begin{equation}
y^s_i=\begin{cases}
1, & \text{if } H > a \\
0, & \text{otherwise}
\end{cases}
\end{equation}

The network is trained to minimize the cross entropy loss between predicted $\hat{y}^s_i$ (softmax of the network output)  and true label $y^s_i$. The basic idea behind this self-supervised framework is that the polarity of events reflects the way objects move in the scene, which could be utilized to derive pre-trained weights for the downstream tasks. As a pre-trained model is learnt from diverse data, it prevents overfitting in the fine-tuning stage while working with limited labelled data.

\section{Experiments and Results}
In this section, we provide comprehensive results on the assessment of the proposed methodology in terms of self-supervised learning and event-only computing. 

\subsection{Dataset and Metrics}
This paper works on two publicly available datasets known as DSEC-Semantic \cite{Messikommer} and DDD17 \cite{ddd17} to evaluate EventSSEG. The driving dataset DDD17 was recorded by the DAVIS346 event camera. DSEC-Semantic, proposed by \cite{Messikommer}, is an extension of the DSEC semantic segmentation dataset made up of 53 driving sequences collected by a high standard frame-based camera and a high-resolution ($640 \times{480}$) monochrome Prophesse Gen3.1 event camera. In this work, we have amalgamated non-road labels into a single class to transform the general semantic segmentation labels into road segmentation task.

\subsection{Comparison of EventSSEG Self-Supervised Learning with Supervised Segmentation Methods}

This section focuses on comparing the efficacy of self-supervised learning of EventSSEG with supervised state-of-the-art event cameras and conventional semantic segmentation networks. In table \ref{table:comparison_msec}, we mainly compare EventSSEG with VID2E \cite{v2e}, EvSegNet \cite{evsegnet}, EvSegFormer \cite{Jia}, LFD$\_$Roadseg \cite{road_new_5}, SegFormer \cite{segformer}, PSPnet \cite{pspnet_1} \cite{pspnet_2}, DeepLabV3 (MobileNetV2) \cite{deeplabv3} \cite{mobilenetv2}, DeepLabV3+ (MobileNetV2) \cite{deeplabv3plus} \cite{mobilenetv2}, FCN (MobileNetV2) \cite{fcn} \cite{mobilenetv2}, MobileNetV3 \cite{mobilenetv3}. These segmentation methods follow the approach of accumulating events into event frames for a window of $50$ms and subsequently applying segmentation networks to process the resultant $2D$ grid. Based on empirical syudy, $N=50$ was found to be optimum for the proposed EventSSEG.

The major challenge of EventSSEG is its learning with limited labelled data. Despite this challenge, the proposed self-supervised task and probabilistic attention mechanism of EventSSEG have paved the way for state-of-the-art accuracy and mean Intersection Over Union (IOU).

\begin{table}
\centering
\begin{tabular}{c|cc||c|cc}
\textbf{Method} & \textbf{Accuracy} & \textbf{mIOU} & \textbf{Method} & \textbf{Accuracy} & \textbf{mIOU}\\ \hline
VID2E  &  $0.94$ & $0.81$ & DeepLabV3+    & $0.94$ & $0.88$\\
EvSegNet  & $0.95$ & $0.79$ & FCN   & $0.89$ & $0.77$ \\ 
EvSegFormer  & $0.95$ & $0.85$ & MobileNetV3  & $0.89$ & $0.79$\\
LFD$\_$Roadseg  & $0.81$ & $0.79$ & PSPnet   & $0.87$ & $0.78$ \\
SegFormer  & $0.94$ & $0.87$ & DeepLabV3   & $0.87$ & $0.77$ \\ \textbf{Proposed}$_a$ & $0.90$ & $0.73$ & \textbf{Proposed}$_b$ & $0.93$ & $0.81$ \\
\hline 
\end{tabular}
\caption{Comparison of EventSSEG with state-of-the-art segmentation methods on DSEC-Semantic dataset. Other segmentation methods are fine-tuned for binary road segmentation tasks with all the labelled training samples (approximately million samples). EventSSEG was initialized with self-supervised pre-trained weights and subsequently fine-tuned with $5.12$k (\textbf{Proposed}$_a$) and $256$k  (\textbf{Proposed}$_b$) samples. }
\label{table:comparison_msec}
\end{table}
	
\subsubsection{Transfer to Other Datasets}
In this section, we delve into evaluating transfer learning performance on the DDD17 dataset in fine-tuning mode. EventSSEG base architecture was trained in self-supervised mode with the DVSEC dataset. Following this pre-training, a dense layer comprising $128$ nodes and ReLU activation and a classification layer with $2$ nodes are appended to the base architecture. During the fine-tuning stage, EventSSEG was trained end-to-end using $256$ labelled samples for $10$ epochs. Optimization was carried out using AdamW optimizer with a learning rate set at $0.001$. The results (Table. \ref{table:comparison_ddd17}) demonstrate that self-supervised EventSSEG performs superior compared to existing supervised semantic segmentation architectures. Note that the proposed EventSSEG was able to achieve state-of-the-art accuracy with limited labelled data of the DDD17 dataset.

\begin{table}
\centering
\begin{tabular}{c|cc}
\textbf{Method} & \textbf{Accuracy} & \textbf{mIOU} \\ \hline
VID2E \cite{v2e} &  $0.96$ & $0.80$ \\
EvSegNet \cite{evsegnet} & $0.96$ & $0.81$ \\ 
EvSegFormer \cite{Jia} & $0.97$ & $0.82$ \\
Mask2Former \cite{mask2former} & $0.92$ & $0.87$ \\
\textbf{Proposed} & $0.95$ & $0.92$ \\ \hline
\end{tabular}
\caption{Comparison of EventSSEG with state-of-the-art segmentation methods on DDD17 dataset. EventSSEG was initialized with self-supervised pre-trained weights (of the DVSEC dataset) and subsequently fine-tuned with $256$k samples. Other segmentation methods were re-trained for the binary road segmentation task with all the labeled samples of the training data.}
\label{table:comparison_ddd17}
\end{table}

\subsection{Comparison of Computational Efficiency}

\begin{table}
\centering
\begin{tabular}{c|ccc}
\hline
\textbf{Method} & \textbf{No of Params (M)} & \textbf{GFLOPs} & \textbf{Time (Secs)} \\ \hline
VID2E \cite{v2e} & $23.75$ & $68.5200$ & $3.950$ \\
EvSegNet \cite{evsegnet} & $23.75$ & $67.5900$ & $1.900$ \\ 
EvSegFormer \cite{Jia} & $24.20$ & $39.3600$ & $2.930$ \\
Mask2Former \cite{mask2former} & $48.70$ & $534.0000$ & $4.476$ \\
\textbf{Proposed} & $\mathbf{3.33}$ & $\mathbf{0.0076}$ & $\mathbf{0.115}$ \\ \hline
\end{tabular}
\caption{Comparison of EventSSEG with state-of-the-art segmentation methods in terms of parameters, GFLOPs and inference time. Note the huge gain achieved by the proposed method in terms of computational efficiency and latency.}
\label{table:comparison_msec_eff}
\end{table}

Table \ref{table:comparison_msec_eff} provides a comparative analysis of the efficiency of EventSSEG with state-of-the-art semantic segmentation networks in terms of a number of parameters, GFLOPs and inference time on NVIDIA GeForce RTX 3090 GPU. Note the huge saving in computational efficiency of EventSSEG.

Event cameras generate sparse and asynchronous events. The proposed processing is designed to utilize these inherent properties of the event camera, aiming to develop a solution that minimizes latency and computational demands. The decision is available at the rate of event occurrence. State-of-the-art event segmentation models accumulate events into event frames and directly apply synchronous frame-based deep learning architectures. This approach imposes a minimum latency determined by the accumulation time.  

In scenarios where the vehicle's movement is slow or the scene contains low texture, the number of events generated is considerably low. The proposed solution, being activity-driven, results in significant power savings during idle situations. Conversely, state-of-the-art event segmentation networks must process a grid of size $N\times{M}$ ($N$ and $M$ are the number of pixels in $x$ and $y$ dierction) even when the number of events occuring is substantially low.

\section{Conclusion}
We introduce a novel event-only transformer-based self-supervised architecture, EventSSEG, for road segmentation using event camera data. By processing raw events directly, it avoids the inefficiencies of dense frame processing. The network is pre-trained in a self-supervised manner with unlabeled events and fine-tuned with limited labeled events. The experiments demonstrated that the proposed approach was able to deliver state-of-the-art performance in terms of accuracy and mean IOU with a minimal number of labelled events. This emphasizes the adaptability of our approach to minimally rely on labelled event data while delivering an efficient event-only road segmentation solution suitable for resource-limited robotics applications.

\end{document}